\documentclass{article}

\usepackage{microtype}
\usepackage{graphicx}
\usepackage{subfigure}
\usepackage{booktabs} 

\usepackage{hyperref}

\usepackage[accepted]{icml2019}

\usepackage{amsmath}

\usepackage{amsfonts}
\usepackage{xcolor}
\usepackage{multirow}

\usepackage{subfigure}

\usepackage{lipsum}
\usepackage{enumitem,calc}

\usepackage{epsfig}
\usepackage{amssymb}

\usepackage{amsfonts}
\usepackage{amsthm}

\newtheorem{thm}{Theorem}
\newtheorem{lem}{Lemma}

\icmltitlerunning{Online Learning and Planning in Partially Observable Domains without Prior Knowledge}

\begin{document}

\twocolumn[
\icmltitle{Online Learning and Planning in Partially Observable Domains without Prior Knowledge}




\begin{icmlauthorlist}
\icmlauthor{Yunlong Liu}{to}
\icmlauthor{Jianyang Zheng}{to}
\end{icmlauthorlist}

\icmlaffiliation{to}{Department of Automation, Xiamen University, Xiamen, China}

\icmlcorrespondingauthor{Yunlong Liu}{ylliu@xmu.edu.cn}

\icmlkeywords{Machine Learning, ICML}

\vskip 0.3in
]



\printAffiliationsAndNotice{}  

\begin{abstract}
How an agent can act optimally in stochastic, partially observable domains is a challenge problem, the standard approach to address this issue is to learn the domain model firstly and then based on the learned model to find the (near) optimal policy. However, offline learning the model often needs to store the entire training data and cannot utilize the data generated in the planning phase. Furthermore, current research usually assumes the learned model is accurate or presupposes knowledge of the nature of the unobservable part of the world. In this paper, for systems with discrete settings, with the benefits of Predictive State Representations~(PSRs), a model-based planning approach is proposed where the learning and planning phases can both be executed online and no prior knowledge of the underlying system is required. Experimental results show compared to the state-of-the-art approaches, our algorithm achieved a high level of performance with no prior knowledge provided, along with theoretical advantages of PSRs. Source code is available at https://github.com/DMU-XMU/PSR-MCTS-Online.
\end{abstract}

\section{Introduction}
\
One commonly used technique for agents operating in stochastic and partially observable domains is to learn or construct an offline model of the underlying system firstly, e.g., Partially Observable Markov Decision Processes~(POMDPs)~\cite{Kaelbling98planningand} or Predictive State Representations~(PSRs)~\cite{Littman01predictiverepresentations,hefny2018recurrent}, then the obtained model can be used for finding the optimal policy. Although POMDPs provide a general framework for modelling and planning in partially observable and stochastic domains, they rely heavily on a prior known and accurate model of the underlying environment~\cite{Ross2008OnlinePOMDP,Silver10monte-carloplanning}. While recent research showed the successful offline PSR-based learning and planning from scratch, offline learning a model needs to store the entire training data set and the model parameters in memory at once and  the data generated during the planning phase is not utilized~\cite{LiuJAIR2019}.

Recently, with the successful applications of online and sample-based Monte-Carlo tree search~(MCTS) techniques, e.g, AlphaGo~\cite{Silver2016}, the BA-POMCP~(Bayes-Adaptive Partially Observable Monte-Carlo Planning) algorithm~\cite{Katt2017,Katt2018}, where the BA-POMDP framework~\cite{Ross2011BAPOMDP} is combined with the MCTS approach, tries to deal with the problem of planning under model uncertainty in larger-scale systems. Unfortunately, the casting of the original problem into an POMDP usually leads to a significant complexity increase~(with intractable large number of possible model states and parameters) and make it even harder to be solved, also, as is well-known, to find an (approximate) optimal solution to an POMDP problem in larger-scale systems is very difficult. Moreover, to guarantee the performance of the POMDP-based approaches, strong prior domain knowledge of the underlying system should be known in advance. 

Unlike the latent-state based approaches, e.g., POMDPs, PSRs work only with the observable quantities, which leads to easier learning of a more expressive model, and more importantly, allows the learning of the model from scratch with no prior knowledge required~\cite{Littman01predictiverepresentations,Liu2015,Liu16IJCAI,JMLR:hamilton14}. In the work of \cite{LiuJAIR2019}, an offline PSR model is firstly learned using training data, then the learned model is combined with MCTS for finding optimal policies. Although compared to the BA-POMDP based approaches, the offline PSR model-based planning approach has achieved significantly better performance and can plan from scratch~\cite{LiuJAIR2019}, as mentioned, offline learning the model often needs to store the entire training data and cannot utilize the data generated in the planning phase, moreover, for the offline learning of the PSR model, $|A|\times|O|$ matrices $P_{T,ao,H}$ are needed to be stored and computed, where $|A|$ the number of actions, $|O|$ the number of observations, $H$ the possible histories and $T$ the possible tests respectively~(See Background for details of history and test). For large scale systems, the storing and calculating such large number of large matrices are expensive and time-consuming~\cite{Boots2011}. Recent research has shown the successful online learning of the PSR model, where the model can be updated incrementally when new training data arrives, and the storing and manipulation of $P_{T,ao,H}$ is not required~\cite{Boots2011,JMLR:hamilton14}. Also, the space complexity of the learned model is independent of the number of training examples and its time complexity is linear in the number of training examples~\cite{Boots2011}.

In this paper, as an alternative and improvement to the offline technique, we introduce an online learning and planning approach for agents operating in partially observable domains with discrete settings, where the model can be learned online and the planning can be started from scratch with no prior knowledge of the domain provided. At each step, the PSR model is online updated and used as the simulator for the computation of the local policies of MCTS. Experiments on some benchmark problems show that with no prior knowledge provided in both cases, our approach performs significantly better than the state-of-the-art BA-POMDP-based algorithms. The effectiveness and scalability of our  approach are also demonstrated by scaling to learn and plan in larger-scale systems, the RockSample problem~\cite{Smith:2004:HSV,Ross2011BAPOMDP}, which is infeasible for the BA-POMDP-based approaches. We further compared our online technique to the offline PSR-based approaches~\cite{LiuJAIR2019}, and experiments show the performance of the online approach is still competitive while requiring less computation and storage resources. As also can be seen from the experiments, with the increase complex of the underlying system, the online approach outperforms the offline techniques while reserving the theoretical advantages of the PSR-related approaches.

\section{Background}
PSRs offer a powerful framework for modelling partially observable systems by representing states using completely observable events~\cite{Littman01predictiverepresentations}. For discrete systems with finite set of observations $O=\{o^1,o^2,\cdots,o^{|O|}\}$ and actions $A=\{a^1,a^2,\cdots,a^{|A|}\}$, at time $\tau$, the observable state representation of the system is a prediction vector of some tests conditioned on current history, where a \emph{test} is a sequence of action-observation pairs that starts from time $\tau + 1$, a \emph{history} at $\tau$ is a sequence of action-observation pairs that starts from the beginning of time and ends at time $\tau$, and the prediction of a length-$m$ test $t$ at history $h$ is defined as $p(t|h)=p(ht)/p(h)=\prod^m_{i=1}Pr(o_i|ha_1o_1\cdots a_i)$~\cite{Singh04predictivestate,LiuJAIR2019}.

The underlying dynamical system can be described by a special bi-infinite matrix, called the Hankel matrix~\cite{Balle:2014ML}, where the rows and columns correspond to the possible tests $\mathcal{T}$ and histories $\mathcal{H}$ respectively, the entries of the matrix are defined as $P_{t,h}=p(ht)$ for any $t \in \mathcal{T}$ and $h \in \mathcal{H}$, where $ht$ is the concatenation of $h$ and $t$~\cite{Boots2011a}. The rank of the Hankel matrix is called the linear dimension of the system~\cite{Singh04predictivestate,Huang2018}. Then a PSR of a system with linear dimension $k$ can be parameterized by a reference condition state vector $b_* = b(\epsilon)\in \mathbb{R}^k$, an update matrix $B_{ao} \in \mathbb{R}^{k \times k}$ for each $a \in A$ and $o \in O$, and a normalization vector $b_\infty \in \mathbb{R}^k$, where $\epsilon$ is the empty history and $b_\infty^T B_{ao} = 1^T$~\cite{Hsu_aspectral,Boots2011a}. Using these parameters, the PSR state at next time step $b(hao)$ can be updated from $b(h)$ as follows~\cite{Boots2011a}:
\begin{small}
\begin{equation}
b(hao) = \frac{B_{ao}b(h)}{b_\infty^TB_{ao}b(h)}.
\label{equ:nextB}
\end{equation}
\end{small}Also, the probability of observing the sequence $a_1o_1a_2o_2 \cdots a_no_n$ in the next $n$ time steps can be predicted by~\cite{Boots2011a,LiuJAIR2019}:
\begin{small}
\begin{equation}
Pr[o_{1:t}||a_{1:t}] = b_{\infty}^TB_{a_no_n} \cdots B_{a_2o_2}B_{a_1o_1}b_*.
\label{equ:nextSeqOb}
\end{equation}
\end{small}
Many approaches have been proposed for the offline PSR model learning\cite{Boots2011a,Liu2015,Liu16IJCAI}, and recently, some online spectral learning approaches of the PSR model have been developed~\cite{Boots2011,JMLR:hamilton14}. The details of those approaches are showed in Appendix A.

It is shown the learned model is consistent with the true model under some conditions~\cite{JMLR:hamilton14,LiuJAIR2019}. Also note rather than offline storing and computing $|A| \times |O|$ matrices $P_{T,ao,H}$ for the offline learning of the PSR model~\cite{LiuJAIR2019}, for the online learning of the PSR model, these matrices are not needed~\cite{Boots2011,JMLR:hamilton14}.
\section{Online Learning and Planning without Prior Knowledge}
For online planning in partially observable and stochastic domains with discrete settings, the most practical solution is to extend MCTS to the model of the underlying environment, where MCTS iteratively builds a search tree with node $T(h)$ that contains $N(h)$, $N(h,a)$, and $V(h,a)$ to combine Monte-Carlo simulation with game tree search~\cite{GellyACM,Silver10monte-carloplanning}. At each decision step, besides for being used for state updating, the model is used to generate simulated experiments for MCTS to construct a lookahead search tree for forming a local approximation to the optimal value function, where each simulation contains two stages: a tree policy and a rollout policy~\cite{Silver10monte-carloplanning,Katt2017}. While current research shows the success of the model-based MCTS on some benchmark problems, either the used model is assumed to be known a prior and accurate~\cite{Silver10monte-carloplanning} or some strong assumptions on the model is made to guarantee the planning performance, e.g, a nearly correct initial model~\cite{Ross2011BAPOMDP,Katt2017,Katt2018} or the model can be only learned offline~\cite{LiuJAIR2019}.

\begin{algorithm}[!htb]
\caption{PSR-MCTS-Online($Z$, $num\_episodes$)}
\label{Algo:PSR Online Learning}
\begin{algorithmic}\tiny
\STATE $//$Get \ initial \  model \ parameters
\STATE ${\textstyle\hat {\sum}_\mathcal H}$, ${\textstyle\hat {\sum}_\mathcal {T,H}}$ $\leftarrow$ Matrices estimated  by initial training data $Z$ in an online mannner\\
\STATE $(\hat U,\hat S,\hat V)= SVD(\textstyle\hat {\sum}_\mathcal {T,H})$
\STATE $\hat b_1$, $\hat b_\infty^T$, $\hat B_{ao}$ $\leftarrow$ Model parameters calculated using $(\hat U,\hat S,\hat V)$
\FOR{$i \leftarrow 1$ {\bfseries to} $n\_episodes$}
\STATE $//$Get new training data $Z^{'}$ and updated model \ parameters
\STATE $Z^{'}$=PSR-MCTS($h$)
\STATE ${\textstyle\hat {\sum}_\mathcal H}$, ${\textstyle\hat {\sum}_\mathcal {T,H}}$ $\leftarrow$ Updated model parameters with new arriving data $Z^{'}$
\STATE $\hat U,\hat S,\hat V$ $\leftarrow$ Execute SVD on updated ${\textstyle\hat {\sum}_\mathcal {T,H}}$ or using the methods
of~\cite{Matthew2002Incremental}
\STATE $\hat b_1$, $\hat B_{ao}$, $\hat b_\infty^T$, $\leftarrow$ Model parameters re-computed in an online manner
\ENDFOR
\end{algorithmic}
\end{algorithm}

\begin{algorithm}[!htb]
\caption{PSR-MCTS($h$)}
\label{Algo:PSR-MCTS}
\begin{algorithmic}\tiny
\STATE $h \leftarrow ()$
\STATE $b(h)\leftarrow \hat{b}_1$
\REPEAT
\STATE $//$Select action according to $\varepsilon$-greedy policy
\STATE $a \leftarrow$ $\left\{\begin{array}{lc}1-\varepsilon&\ \ Act-Search(b\left(h\right),n\_sims,h)\\\frac\varepsilon{\vert A\vert-1}&for\;all\;other\; |A|-1 \;actions\;a\end{array}\right.$
\STATE
\STATE{\bfseries EXECUTE} $a$
\STATE $o \leftarrow$ observation received from the world
\STATE $//$Update belief state based on received observations
\STATE $b(hao) = \frac{\hat{B}_{ao}b(h)}{\hat{b}_\infty^T\hat{B}_{ao}b(h)}$
\STATE $h \leftarrow hao$
\UNTIL{the end of a plan}
\STATE{\bfseries return} $h$
\end{algorithmic}
\end{algorithm}

In this section, we show how the online-learned PSR model can be combined with MCTS. As mentioned, PSRs use completely observable quantities for the state representation and the spectral learned PSR model is consistent with the true model, moreover no prior knowledge is required to learn a PSR model~\cite{JMLR:hamilton14}. In our approach, with the benefits of PSRs and its online learning approach, the PSR model is firstly learned and updated in an online manner~(Algorithm~\ref{Algo:PSR Online Learning}). Then at each decision step, as commonly used in the literature~\cite{Silver10monte-carloplanning,Katt2017,LiuJAIR2019}, the (learned) model can be straightforwardly used to generate the simulated experiments for MCTS to find the good local policy~(Note that functions Act-Search and Simulate are the same as in the work of \cite{LiuJAIR2019}, we refer the readers to see it or Appendix B for the detail). After the found action is executed and a real observation is received, the model and the next state representation is updated. For the MCTS process, in the simulation, at each time step, after an action is selected, an observation is sampled according to $Pr[o||ha] = \hat{b}_{\infty}^T\hat{B}_{ao}b(h)$ by following Equ.\ref{equ:nextSeqOb}.

Then the state in the simulation is also updated. Note that in the POMDP-based MCTS approaches, such state update and observation sampling are also needed~\cite{Silver10monte-carloplanning,Katt2017,Katt2018}, and as shown in the work of \cite{LiuJAIR2019}, when compared to the POMDP-based approaches, the state representation of the PSR-based approach is more compact and the computation of the next observation is more efficient.

In the partially observable environments, as we cannot know exactly some states, it is difficult to determine the rewards related to these states. In our approach, as used in the offline technique~\cite{LiuJAIR2019}, we directly treat some rewards as observations, e.g., the goal related rewards, and these observations~(rewards) are used in the online model learning. In the simulation, when the sampled observation $o$ is the reward that indicates the end of a process, the simulation ends. Otherwise, the simulation ends with some pre-defined conditions. Note that some state-independent rewards, such as the rewards received at every time step or action-only-dependent rewards in some domains, are not treated as observations and not used for the model learning~\cite{LiuJAIR2019}. After a real observation $o$ is received from the world, $h \leftarrow hao$, $b(h)$ is updated according to Equ.~\ref{equ:nextB}, and the node $T(h)$ becomes the root of the new search tree. 

\section{Experimental Results}
\subsection{Experimental setting}

\begin{figure*}[!ht]
\begin{minipage}{7in}
\centerline{\hspace{0.5in}{\scriptsize {\em Tiger}}\hspace{1.3in}{\scriptsize {\em POSyadmin}}\hspace{1.1in}{\scriptsize {\em RockSample(5,5)}}\hspace{1.0in}{\scriptsize {\em RockSample(5,7)}}}
\begin{minipage}{1.6in}
\includegraphics[width=1.0\textwidth]{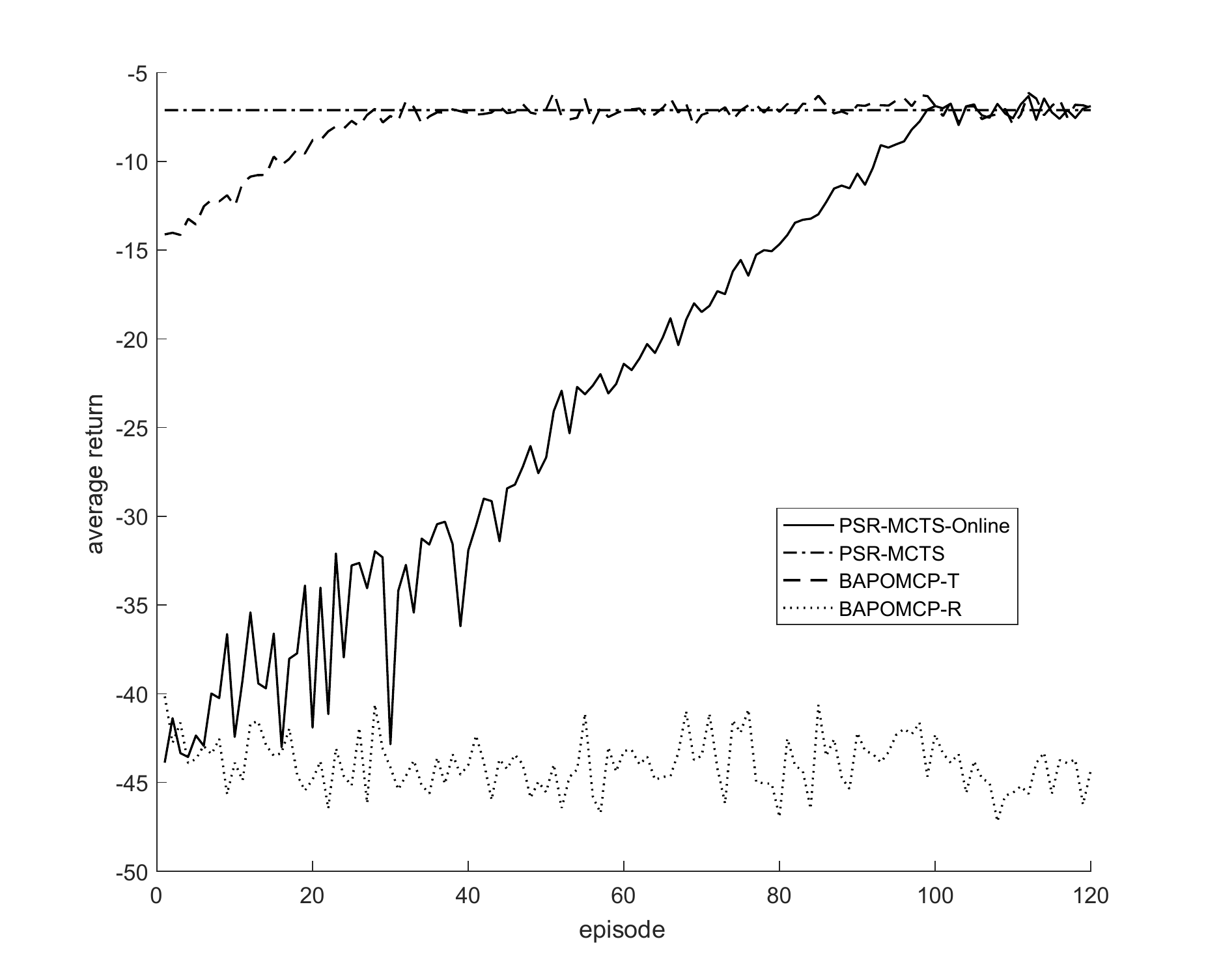}
\centerline{{\small ($a$)}}
\end{minipage}
\begin{minipage}{1.6in}
\includegraphics[width=1.0\textwidth]{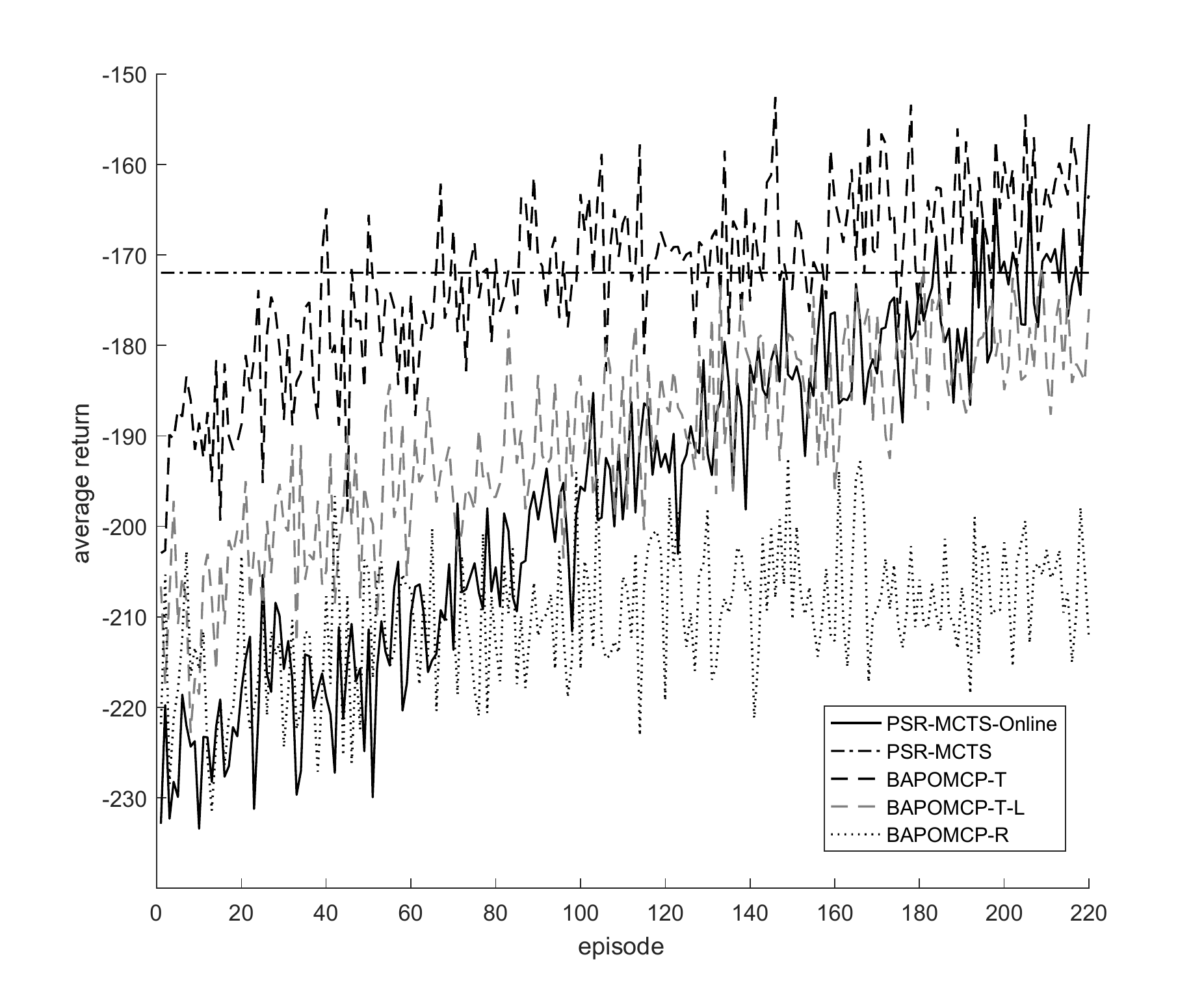}
\centerline{{\small ($b$)}}
\end{minipage}
\begin{minipage}{1.6in}
\includegraphics[width=1.0\textwidth]{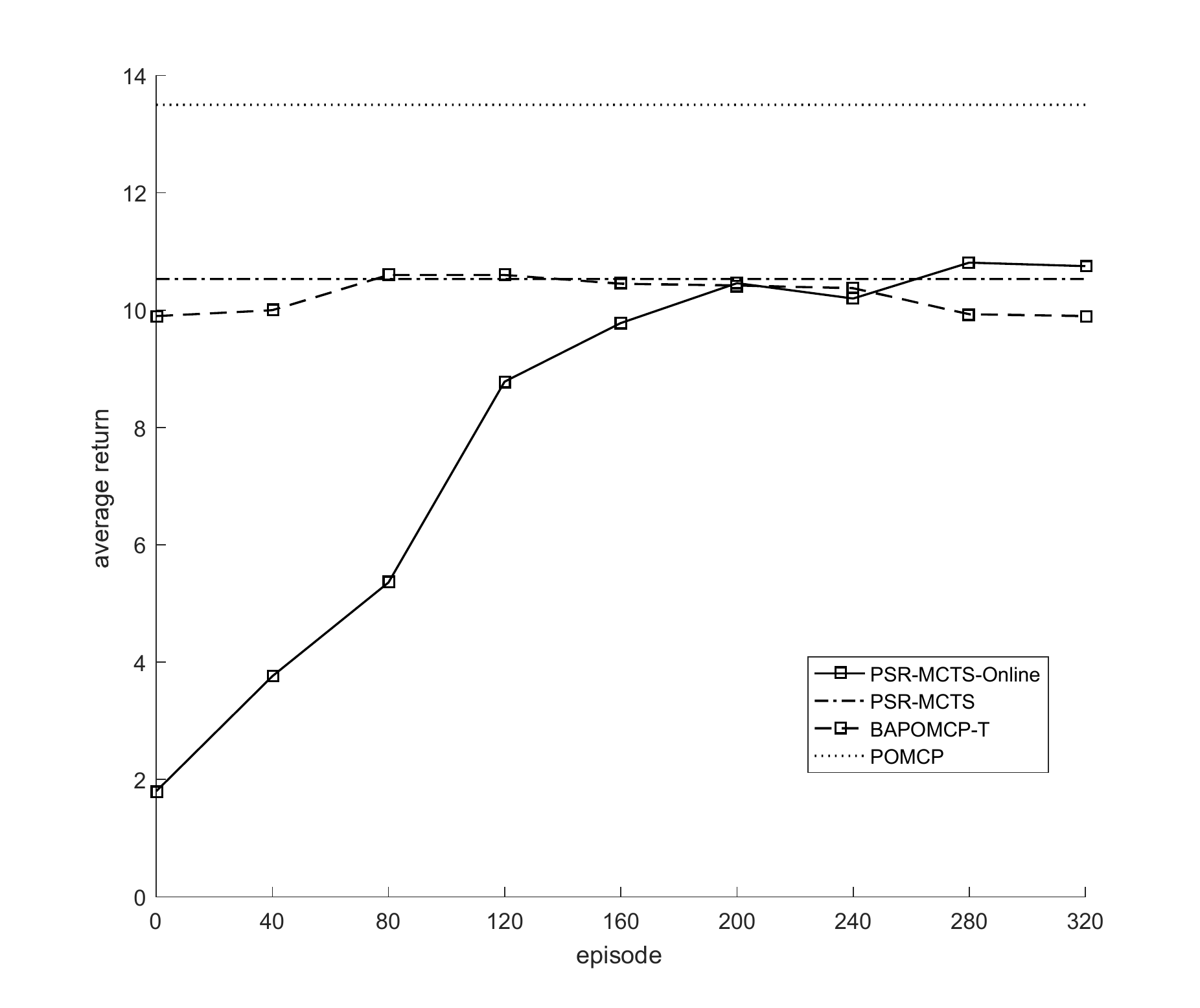}
\centerline{{\small ($c$)}}
\end{minipage}
\begin{minipage}{1.6in}
\includegraphics[width=1.0\textwidth]{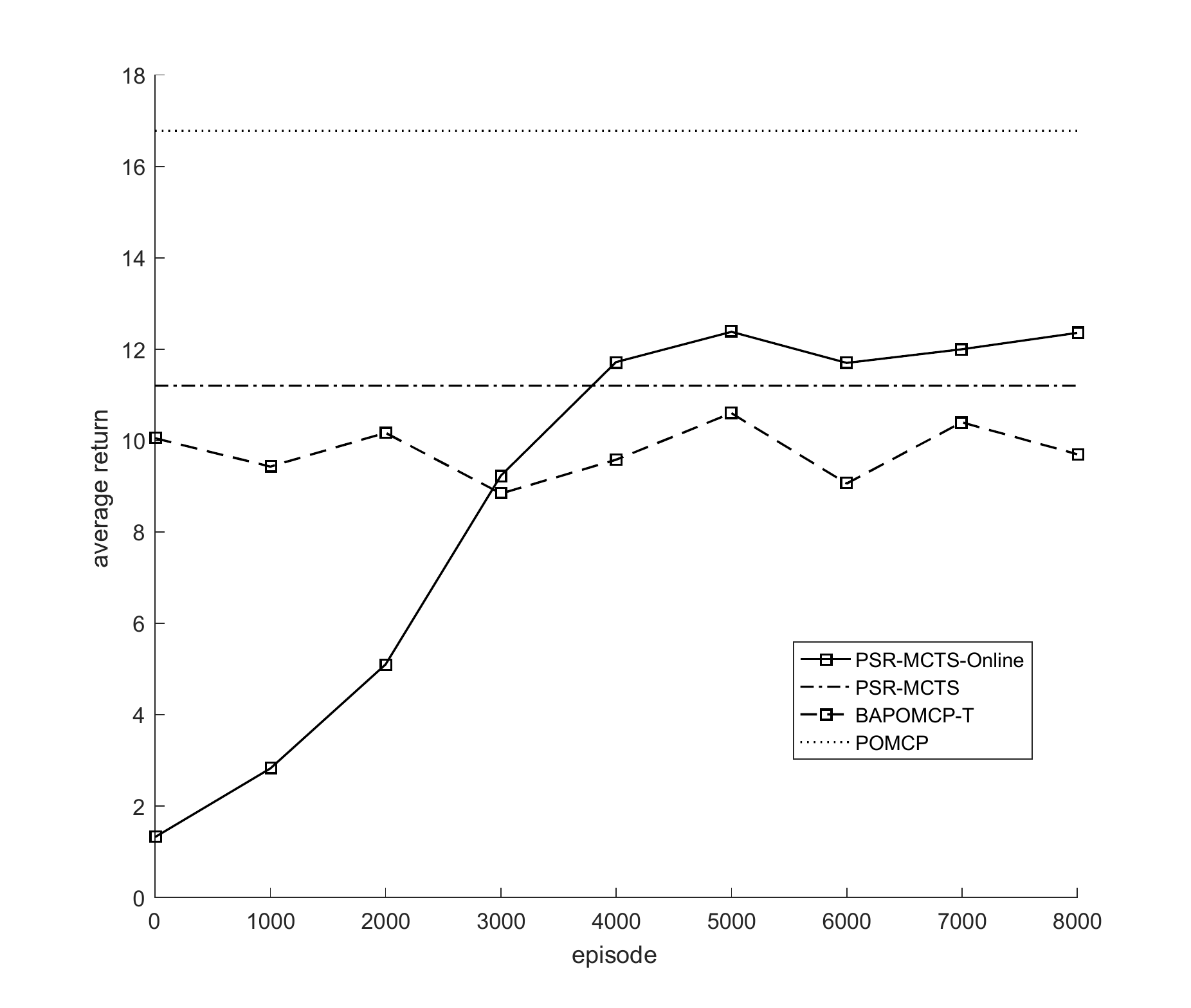}
\centerline{{\small ($d$)}}
\end{minipage}
\caption{{\scriptsize Average return for different algorithms in different domains}}
\end{minipage}
\end{figure*}

To evaluate our proposed approach, we first executed our algorithm on Tiger and POSyadmin, the two problems used to test the state-of-the-art BA-POMCP approach~\cite{Katt2017,Katt2018}. Then we extend our approach to RockSample(5,5) and RockSample(5,7)~\cite{Smith:2004:HSV,Ross2011BAPOMDP}, both of which are too complex for the BA-POMDP based approaches.

The details of the experimental settings in our approach are showed in Appendix C.

\subsection{Evaluated methods}

As mentioned, the performance of the BA-POMDPs based approaches rely heavily on knowing the knowledge of the underlying system~\cite{Katt2017,Katt2018}, for example, in the work of \cite{Katt2017}, for Tiger, the transition model is assumed to be correct and the initial observation function is assumed to be nearly correct; for POSyadmin, the observation function is assumed known a prior, and the initial transition function is assumed to be nearly correct; for RockSample, in the work of~\cite{Ross2011BAPOMDP}, the transition model is assumed to be correct and the initial observation function is assumed to be nearly correct.

To evaluate our method, for the first two problems, we firstly compared our approach to the BA-POMCP method under the same conditions~(BAPOMCP-R), that is, no prior knowledge is provided to the BA-POMCP approach, then our approach was compared to the BA-POMCP approach with the nearly correct initial models~(BAPOMCP-T) as mentioned previously. For POSyadmin, we also compared our approach to the BA-POMCP approach with the less accurate initial model~(BAPOMCP-T-L), where the true probabilities of transition function are either subtracted or added by $0.4$, and for probabilities fall below $0$, they are set to $0.001$. Then, each Dirichlet distribution is normalized with the counts summing to $20$. For the RockSample problem, our approach was compared to the BA-POMCP approach with the nearly correct initial models~(BAPOMCP-T). We also show the result of the POMCP approach~\cite{Silver10monte-carloplanning} where the accurate models of the underlying systems are used. To further verify the performance of the online approach, we also compared with the offline PSR model learning-based approach~(PSR-MCTS)~\cite{LiuJAIR2019} where the PSR model is firstly learned offline and then combined with MCTS for planning.

\subsection{Performance evaluation}

Figure 1:($a$) plots the average return over 10000 runs with 10000 simulations at each decision step of MCTS on Tiger. For Tiger, for the BAPOMCP-R approach, nearly no improvement has been achieved with the increase of episodes and for the BAPOMCP-T approach, with the increase of the episodes, the performance becomes stable. For our PSR-MCTS-Online approach, with the increase of the learning episodes, our approach performs significantly better than the BAPOMCP-R approach. And finally, our approach achieves the same performance with the BAPOMCP-T approach, which has strong prior knowledge provided.

Experimental results for four approaches on the (3-computer) POSysadmin problem are shown in Figure 1:($b$), where 100 simulations per step were used. For the BAPOMCP-T and BAPOMCP-T-L approach, the results tend to be stable after 150 episodes, and for the BAPOMCP-R approach, nearly no improvement has been achieved with the increase of the episodes. Our PSR-MCTS-Online approach still performs significantly better than the BAPOMCP-R approach and finally achieves nearly the same performance of the BAPOMCP-T approach. While for the BAPOMCP-T-L approach, after less than $150$ episodes, our approach has a better performance.

Figure 1:($c$) and Figure 1:($d$) plot the average return over 1000 runs with 1000 simulations on RockSample(5,5) and RockSample(5,7). For such scale systems, the state size of the BA-POMDP model is intractable large with the increase of the time step, and we may not be able even to store and initialize the state transaction matrices. For the convenience of the comparison, as used in the work of \cite{Ross2011BAPOMDP}, only the dynamics related to the check action were modeled via the BA-POMDP approach, for the others, the black box simulation of the exact model was used to generate the simulated experiments and for state representation and updating~(BAPOMCP-T). Even under such conditions, the PSR-MCTS-Online method without no prior knowledge provided finally still achieved the nearly same or better performance with the BAPOMDP-T approach. Even compared to the POMCP method where the underlying model is accurate, the performance of our approach is still competitive. 

When compared with the offline based approach~\cite{LiuJAIR2019}, while without requiring the storing and computation of the $|A| \times |O|$ $P_{T,ao,H}$ matrices, for Tiger, POSysadmin and RockSample(5,5), the online approach eventually achieved the same performance of the offline method, and with the increase of the complexity of the underlying domain, for RockSample(5,7), with the increase of the episodes, the online approach outperforms the offline technique.

\section{Conclusion}
In this paper, we presented PSR-MCTS-Online, a method that can simultaneously learn and plan in an online manner while not requiring any prior knowledge provided. Experimental results show the effectiveness and scalability of the proposed method with the theoretical advantages of the PSR-related approaches. When compared to the offline approach, the effectiveness of the online approach is also demonstrated while requiring less computation and storage resource.

\section*{Acknowledgements}

This work was supported by the National Natural Science Foundation of China (No.61772438 and No.61375077).

\nocite{langley00}

\bibliography{example_paper}
\bibliographystyle{icml2019}

\renewcommand\thesection{\Alph{section}}
\renewcommand\thesubsection{\thesection.\arabic{subsection}}

\pagestyle{fancy} \chead{Online Learning and Planning in Partially Observable Domains without Prior Knowledge}

\twocolumn[
\icmltitle{Supplementary Material: Online Learning and Planning in Partially Observable Domains without Prior Knowledge}




\icmlsetsymbol{equal}{*}
\icmlkeywords{Machine Learning, ICML}

\vskip 0.3in
]




\setcounter{section}{0}
\section{Online Spectral Algorithm}
 Given any length training trajectory $z$ of action-observation pairs in a batch of training trajectories $Z$, three indicator functions, $\mathbb I_{h_j(z)}$, $\mathbb I_{h_j,t_i,(z)}$ and $\mathbb I_{h_j,ao,t_i,(z)}$ for $z\in Z$, are firstly defined, where $\mathbb I_{h_j(z)}$ takes a value of 1 if the action-observation pairs in $z$ correspond to $h_j$, $\mathbb I_{h_j,t_i,(z)}$ takes a value of 1 if $z$ can be partitioned to make that there are $\left|h_j\right|$ action-observation pairs corresponding to those in $h_j\in \mathcal H$ and the next $\left|t_i\right|$ pairs correspond to those in $t_i\in \mathcal T$, and $\mathbb I_{h_j,ao,t_i,(z)}$  takes a value of 1 if $z$ can be partitioned to make that there are $\left|h_j\right|+1$ action-observation pairs corresponding to $h_j$ appended with a particular $ao\in A\times O$ and the next $\left|t_i\right|$ pairs correspond to those in $t_i\in \mathcal T$ respectively~\cite{JMLR:hamilton14}. Then the following matrices can be estimated:
\begin{equation}
{\textstyle\hat {\sum}_\mathcal H}=\sum_{z\in Z}\mathbb I_{h_j(z)}\\
\label{equ:sumH}
\end{equation}

\begin{equation}
{\textstyle\hat {\sum}_\mathcal {T,H}}=\sum_{z\in Z}\;\sum_{t_i,h_j\in \mathcal T\times \mathcal H}\mathbb I_{h_j,t_i,(z)}\\
\label{equ:sumTH}
\end{equation}
Next, an SVD is executed on matrix $\textstyle\hat {\sum}_\mathcal {T,H}$ to obtain $(\hat U,\hat S,\hat V)= SVD(\textstyle\hat {\sum}_\mathcal {T,H})$, then the corresponding model parameters are as follows:

\begin{equation}
\hat b_1 = \hat{S} \hat{V}^Te
\label{equ:b1}
\end{equation}

\begin{equation}
\hat b_\infty^T=\hat{\textstyle\sum}_H^T\hat V\hat S^{-1}
\label{equ:binf}
\end{equation}

and for each $ao \in A \times O$:
\begin{equation}
\hat B_{ao}=\sum_{z\in Z}\;\sum_{t_i,h_j\in \mathcal T\times \mathcal H}\mathbb I_{h_j,ao,t_i}\left(z\right)\left[\hat U^T\left(t_i\right)\hat S^{-1}\hat V^T\left(h_j\right)\right]
\label{equ:Bao}
\end{equation}

where $e$ is a vector such that $e=(1,0,0,\cdots,0)^T$. This specification of $e$ assumes all $z\in Z$ are starting from a unique start state. If this is not the case, then $e$ is set such that $e=(1,1,1,\cdots,1)^T$. In this latter scenario, rather than learning $\hat b_1$ exactly, $\hat b_*$ is learned, which is an arbitrary feasible state as the start state~\cite{JMLR:hamilton14}. $\hat U^T\left(t_i\right)$ is the $t_i^{th}$ row of matrix $\hat U^T$, and $\hat S^{-1}\hat V^T$$\left(h_j\right)$ is the $h_j^{th}$ column of matrix $\hat S^{-1}\hat V^T$~\cite{Boots2011a,JMLR:hamilton14}.

With the arriving of new training data $Z^{'}$, the model parameters can be incrementally updated in an online manner that ${\textstyle\hat {\sum}_\mathcal H}$ and ${\textstyle\hat {\sum}_\mathcal {T,H}}$ are updated accordingly by using Equ.~\ref{equ:sumH} and \ref{equ:sumTH}, then a new $(\hat U_{new},\hat S_{new},\hat V_{new})$ can be obtained by executing SVD on the updated ${\textstyle\hat {\sum}_\mathcal {T,H}}$. Note that in the case of a large amount of training data, the method proposed in the work of~\cite{Matthew2002Incremental} can be used for the updating of ${\textstyle\hat {\sum}_\mathcal {T,H}}$ for improving the efficiency of the calculation. Accordingly, parameters $\hat b_{1}$ and $\hat b_\infty$ can be re-computed by using Equ.~\ref{equ:b1} and Equ.~\ref{equ:binf}, $\hat B_{ao}$ can be updated as follows with $Cnt_{ao}=\sum_{z\in Z}\;\sum_{t_i,h_j\in \mathcal T\times \mathcal H}\mathbb{I}_{h_j,ao,t_i}\left(z\right)$:

\begin{equation}
\begin{split}
&\hat B_{ao}=\\
&\sum_{t_i,h_j\in \mathcal T\times \mathcal H}Cn{t_{ao}}\left[t_i,h_j\right]{\hat U^T}_{new}\left(t_i\right)\hat S^{-1}_{new}{\hat V^T}_{new}\left(h_j\right)
\end{split}
\label{equ:Bao1}
\end{equation}

When the total amount of training data is large, $\hat{B}_{ao}$ can also be updated directly as follows~\cite{JMLR:hamilton14}:
\begin{equation}
\begin{split}
&\hat B_{ao}=\\
&\sum_{z\in Z^{'}}\;\sum_{t_i,h_j\in \mathcal T\times \mathcal H}\mathbb{I}_{h_j,ao,t_i}\left(z\right)\left[\hat U_{new}^T\left(t_i\right){\hat S_{new}}^{-1}\hat V_{new}^T\left(h_j\right)\right]+\\
&\hat U_{new}^T{\hat U}_{old}\hat B_{ao}^{old}{\hat S}_{old}\hat V_{old}^T{\hat V}_{new}\hat S_{new}^{-1}
\end{split}
\label{equ:Bao2}
\end{equation}
It is shown the learned model is consistent with the true model under some conditions~\cite{JMLR:hamilton14,LiuJAIR2019}. Also note that rather than offline storing and computing $|A| \times |O|$ matrices $P_{T,ao,H}$ for the offline learning of the PSR model as mentioned previously, for the online learning of the PSR model, these matrices are not needed.

\section{PSR-MCTS-Online Algorithm}

In the algorithms, $reward(ao)$ is the reward that is not treated as observation, $\gamma$ is a discounted factor specified by the environment, $n\_sims$ is the number of simulations used for MCTS to find the executed action at each step, $max\_dep$ and $IsTerminal(h)$ are some predefined conditions for the termination~(Note that Functions Act-Search, Simulate and RollOut are the same as in the work of \cite{LiuJAIR2019}).

\begin{algorithm}[!htb]
\caption{PSR-MCTS-Online($Z$, $num\_episodes$)}
\label{Algo:PSR Online Learning}

\begin{algorithmic}\tiny
\STATE $//$Get \ initial \  model \ parameters
\STATE ${\textstyle\hat {\sum}_\mathcal H}$, ${\textstyle\hat {\sum}_\mathcal {T,H}}$ $\leftarrow$ Estimated  by initial training data $Z$ using Equ.~\ref{equ:sumH} \\
\ \ \ \ \ \ \ \ \ \ \ \ \ \ \ \ \ \ \ \ \ \ \ \ \ \ and Equ.~\ref{equ:sumTH}
\STATE $(\hat U,\hat S,\hat V)= SVD(\textstyle\hat {\sum}_\mathcal {T,H})$
\STATE $\hat b_1$, $\hat b_\infty^T$, $\hat B_{ao}$ $\leftarrow$ Calculated by Equ.~\ref{equ:b1}, Equ.~\ref{equ:binf} and Equ.~\ref{equ:Bao}
\FOR{$i \leftarrow 1$ {\bfseries to} $n\_episodes$}
\STATE $//$Get new training data $Z^{'}$ and updated model \ parameters
\STATE $Z^{'}$=PSR-MCTS($h$)
\STATE ${\textstyle\hat {\sum}_\mathcal H}$, ${\textstyle\hat {\sum}_\mathcal {T,H}}$ $\leftarrow$ Updated by Equ.~\ref{equ:sumH} and Equ.~\ref{equ:sumTH} with $Z^{'}$
\STATE $\hat U,\hat S,\hat V$ $\leftarrow$ Execute SVD on updated ${\textstyle\hat {\sum}_\mathcal {T,H}}$ or using the methods
of~\cite{Matthew2002Incremental}
\STATE $\hat b_1$, $\hat b_\infty^T$, $\leftarrow$ Re-computed by using Equ.~\ref{equ:b1}, Equ.~\ref{equ:binf}
\STATE $\hat B_{ao}$ $\leftarrow$ Updated by using Equ.~\ref{equ:Bao1} or Equ.~\ref{equ:Bao2}
\ENDFOR
\end{algorithmic}
\end{algorithm}

\begin{algorithm}[!htb]
\caption{PSR-MCTS($h$)}
\label{Algo:PSR-MCTS}
\begin{algorithmic}\tiny
\STATE $h \leftarrow ()$
\STATE $b(h)\leftarrow \hat{b}_1$
\REPEAT
\STATE $//$Select action according to $\varepsilon$-greedy policy
\STATE $a \leftarrow$ $\left\{\begin{array}{lc}1-\varepsilon&\ \ Act-Search(b\left(h\right),n\_sims,h)\\\frac\varepsilon{\vert A\vert-1}&for\;all\;other\; |A|-1 \;actions\;a\end{array}\right.$
\STATE
\STATE{\bfseries EXECUTE} $a$
\STATE $o \leftarrow$ observation received from the world
\STATE $//$Update belief state based on received observations
\STATE $b(hao) = \frac{\hat{B}_{ao}b(h)}{\hat{b}_\infty^T\hat{B}_{ao}b(h)}$
\STATE $h \leftarrow hao$
\UNTIL{the end of a plan}
\STATE{\bfseries return} $h$

\end{algorithmic}
\end{algorithm}

\begin{algorithm}[!htb]
\caption{Act-Search($b(h)$,$n\_sims$,$h$)}
\label{Algo:Action-Search}
\begin{algorithmic}\tiny
\STATE $h_0 \leftarrow h$
\STATE $\bar{b}(h_0) \leftarrow Copy(b(h))$
\FOR{$i \leftarrow 1$ {\bfseries to} $n\_sims$}
\STATE $Simulate(\bar{b}(h_0),0,h_0)$
\ENDFOR
\STATE $a \leftarrow GreedyActionSelection(h_0)$
\STATE{\bfseries return} $a$

\end{algorithmic}
\end{algorithm}

\begin{algorithm}[tb]
\caption{Simulate($b(h),depth,h$)}
\label{Algo:Simulate}
\begin{algorithmic}\tiny
\IF {$depth == max\_dep \| IsTerminal(h)$}
\STATE {\bfseries return} $0$
\ENDIF
\STATE $//$Select action according to the UCT algorithm~\cite{Kocsis:2006}
\STATE $a \leftarrow$ UCBACTIONSELECTION($h$)
\STATE $//$Obtain observation using the PSR method
\STATE $o \leftarrow$ sampled according to $Pr[o||ha] = \hat{b}_{\infty}^T\hat{B}_{ao}b(h)$
\IF{$o$ corresponds to some reward}
\STATE $R \leftarrow reward(o)$
\ELSE
\STATE $R \leftarrow reward(ao)$
\ENDIF
\STATE $//$Update belief state
\STATE $h' \leftarrow hao$
\STATE $b(h') = \frac{\hat{B}_{ao}b(h)}{\hat{b}_\infty^T\hat{B}_{ao}b(h)}$
\IF{$h' \in Tree$}
\STATE $r \leftarrow R+\gamma\cdot$Simulate($b(h'),depth+1,h'$)
\ELSE
\STATE ConstructNode($h'$)
\STATE $r \leftarrow R+\gamma\cdot$RollOut($b(h'),depth+1,h'$)
\ENDIF

\STATE $//$Update statistics. $N(h)$: number of times history $h$ has been visited. $V(h,a)$: value of $h$
\STATE $N(h) \leftarrow N(h)+1$
\STATE $N(h,a) \leftarrow N(h,a)+1$
\STATE $V(h,a) \leftarrow V(h,a)+\frac{r-V(ha)}{N(ha)}$
\STATE{\bfseries return $r$}

\end{algorithmic}
\end{algorithm}

\begin{algorithm}[!tb]
\caption{RollOut($b(h),depth,h$)}
\label{Algo:RollOut}
\begin{algorithmic}\tiny
\IF {$depth == max\_dep \| IsTerminal(h)$}
\STATE {\bfseries return} $0$
\ENDIF
\STATE $//$Using the rollout policy (random policy) to select the action to perform
\STATE $a \leftarrow \pi_{rollout}(h)$
\STATE $//$Obtain observation using the PSR method
\STATE $o \leftarrow$ sampled according to $Pr[o||ha] = \hat{b}_{\infty}^T\hat{B}_{ao}b(h)$
\IF{$o$ corresponds to some reward}
\STATE $R \leftarrow reward(o)$
\ELSE
\STATE $R \leftarrow reward(ao)$
\ENDIF
\STATE $//$Update belief state
\STATE $h' \leftarrow hao$
\STATE $b(h') = \frac{\hat{B}_{ao}b(h)}{\hat{b}_\infty^T\hat{B}_{ao}b(h)}$
\STATE $r \leftarrow R+\gamma\cdot$RollOut($b(h'),depth+1,h'$)
\STATE{\bfseries return} $r$

\end{algorithmic}
\end{algorithm}

\section{Experimental setting}

\begin{figure}[!ht]
\begin{minipage}{3.5in}
\centering
\includegraphics[width=0.8\textwidth]{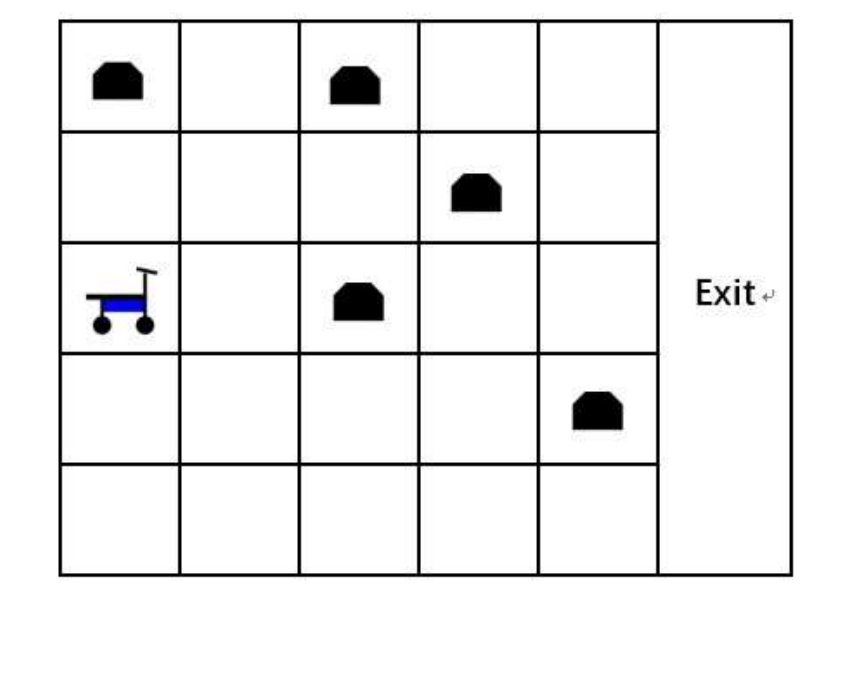}
\caption{{\small {\em RockSample(5,5)}.}}
\label{fig:Rocksample5}
\end{minipage}
\end{figure}

\begin{figure}[!ht]
\begin{minipage}{3.5in}
\centering
\includegraphics[width=0.8\textwidth]{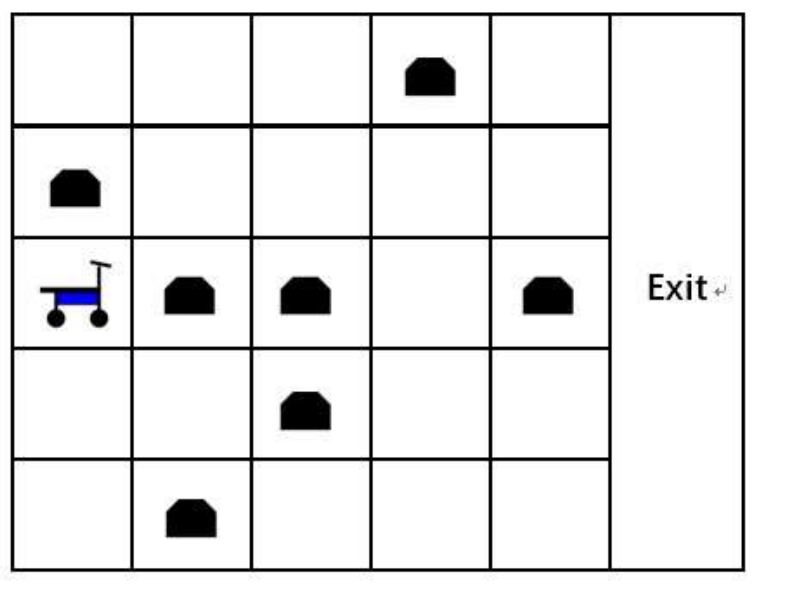}
\caption{{\small {\em RockSample(5,7)}.}}
\label{fig:Rocksample7}
\end{minipage}
\end{figure}

For POSyadmin, the agent acts as a system administrator to maintain a network of $n$ computers. The agent doesn't know the state of any computer, and at each time step, any of the computers can 'fail' with some probability $f$~\cite{Katt2017}. In the RockSample($n,k$) domain~(Figure~\ref{fig:Rocksample5} and \ref{fig:Rocksample7}), a robot is on an $n \times n$ square board, with $k$ rocks on some of the cells. Each rock has an unknown binary quality~(good or bad). The goal of the robot is to gather samples of the good rocks. The state of the robot is defined by the position of the robot on the board and the quality of all the rocks and there is an additional terminal state, reached when the robot moves into exit area, then with a $n \times n$ board and $k$ rocks, the number of states is $n^22^k+1$~\cite{Ross2011BAPOMDP}.

For Tiger, besides the two observations of the original domain, the reward 10 for opening the correct door and the penalty $-100$ for choosing the door with the tiger behind it are treated as observations. For POSyadmin, we added the rewards that indicate the whole status of the network as observations, which provides the information about how many computers have been failed at current step, but we still don't know which computer has been failed, the same rewards were also used in the BA-POMCP based experiments. For RockSample, we treat the reward $10$ for sampling a good rock or moving into exit area and the penalty $-10$ for sampling a bad rock as observations. Also, as the PSR model is learned from scratch, the online learning and planning of our approach is divided into two phases. In the first phase, no model-based decision has been made, and the random policy is used to explore the environment to obtain an initial model. Then, while the model is still updated with the arriving of new training data, the $\varepsilon$-greedy policy is used in the decision process, where $\varepsilon$ will decrease to zero with the increase of the episodes.

The details of the experimental settings in our approach are as follows: For Tiger, the maximum length of $\mathcal{H}$ is set to $6$ and the maximum length of $\mathcal{T}$ is set to $2$; For POSyadmin, the maximum length of $\mathcal{H}$ is set to $12$ and the maximum length of $\mathcal{T}$ is set to $1$; For RockSample(5,5) and RockSample(5,7), the maximum length of $\mathcal{H}$ is set to $27$ and the maximum length of $\mathcal{T}$ is set to $2$.  A random strategy is used in the first 20, 20, 40, and 1000 episodes for Tiger, POSyadmin, RockSample(5,5) and RockSample(5,7) respectively. For RockSample, the random policy is as follows: Each rock is checked twice at most; when the robot is in the cell of the rock, once the rock has been checked, the sample action will be performed; otherwise the sample action will be performed with a 50\% probability. In all other cases, the robot randomly selects an action. For tiger, $\varepsilon$ decreases from $0.5$ to $0$ at a constant speed from the 21st episode to the 100th episode. For POSyadmin, $\varepsilon$ decreases from $0.85$ to $0$ at a constant speed from the 21st episode to the 200th episode. For RockSample(5,5), from the 41st episode to the 200th episode, $\varepsilon$ is initially set to 0.8, then at every 40 episodes $\varepsilon$ is reduced by 0.2.  For RockSample(5,7), from the 1001st episode to the 5000th episode, $\varepsilon$ is initially set to 0.8, and at every 1000 episodes $\varepsilon$ is reduced by 0.2. Note that some different parameters have also been tested, and similar performance has been achieved for our approach.

\section{Related Works}

POMDPs are a general solution for the modelling of controlled systems, however, it is known that learning offline POMDP models using some EM-like methods is very difficult and suffers from local minima, moreover, these approaches presuppose knowledge of the nature of the unobservable part of the world. As an alternative, PSRs provide a powerful framework by only using observable quantities. Much effort has been devoted to learning offline PSR models. In the work of Boots {\em et al.}~\cite{Boots2011a}, the offline PSR model is learned by using spectral approaches. Hamilton {\em et al.}~\cite{JMLR:hamilton14} presented the compressed PSR~(CPSR) models, and the technique learns approximate PSR models by exploiting a particularly sparse structure presented in some domains. The learned CPSR model is also combined with Fitted-Q for planning, but prior knowledge, e.g., domain knowledge, is still required. Hefny et al. \cite{hefny2018recurrent} proposed
Recurrent Predictive State Policy (RPSP) networks, which consist of a recursive
filter for the tracking of a belief about the state of the environment, and a reactive
policy that directly maps beliefs to actions, to maximize the cumulative reward. In the work of \cite{LiuJAIR2019}, an offline PSR model is firstly learned using training data, then the learned model is combined with MCTS for finding optimal policies. Although compared to the BA-POMDP based approaches, the offline PSR model-based planning approach has achieved significantly better better performance, offline learning the model often needs to store the entire training data and cannot utilize the data generated in the planning phase, moreover, for the offline learning of the PSR model, $|A|\times|O|$ matrices $P_{T,ao,H}$.

To alleviate the shortcoming of offline PSR model learning techniques that the entire training data is needed to store in memory at once, the online PSR model learning approach has been proposed in the work of \cite{Boots2011,JMLR:hamilton14}, where the model can be updated incrementally when new training data arrives, also, the learned model's space complexity is independent of the number of training examples and its time complexity is linear in the number of training examples.

When the model of the underlying system is available, model-based planning approaches offer a principled framework for solving the problem of choosing optimal actions, however, most of the related methods assume an accurate model of the underlying system to be known a prior, which is usually unrealistic in real-world applications. The BA-POMDP approach tackles this problem by using a Bayesian approach to model the distribution of all possible models and allows the models to be learned during execution~\cite{Ross2011BAPOMDP,Ghavamzadeh2015BRLSurvey}. Unfortunately, BA-POMDPs are limited to some trivial problems as the size of the state space over all possible models is too large to be tractable for non-trivial problems.

With the benefits of online and sample-based planning for solving larger problems~\cite{Kearns2002,Ross2008OnlinePOMDP,Silver10monte-carloplanning}, some approaches have been proposed to solve the BA-POMDP model in an online manner. In the work of~\cite{Ross2011BAPOMDP}, an online POMDP solver is proposed by focusing on finding the optimal action to perform in the current belief of the agent. Katt et al.~\cite{Katt2017,Katt2018} extend the Monte-Carlo Tree Search method POMCP to BA-POMDPs, results in the state-of-the-art framework for learning and planning in BA-POMDPs. In the work of \cite{Katt2018}, a Factored Bayes-Adaptive POMDP
model is introduced by exploiting the underlying structure of some specific domains. While these approaches show promising performance on some problems, like other Bayesian-based approaches in the literature, the performance is very dependent on the prior knowledge.

\nocite{langley00}

\bibliography{example_paper}
\bibliographystyle{icml2019}

\end{document}